\documentclass{article}

\usepackage[final]{nips_2016}

\usepackage[utf8]{inputenc} 
\usepackage[T1]{fontenc}    
\usepackage{hyperref}       
\usepackage{url}            
\usepackage{booktabs}       
\usepackage{amsfonts}       
\usepackage{nicefrac}       
\usepackage{microtype}      
\usepackage{graphicx}
\usepackage{wrapfig}

\title{Unsupervised Typography Transfer}

\author{
  Hanfei Sun\\
  Carnegie Mellon University\\
  \texttt{hanfeis@andrew.cmu.edu} \\
  \And
  Yiming Luo \\
  Carnegie Mellon University \\
  \texttt{yimingl1@andrew.cmu.edu} \\
  \And
  Ziang Lu \\
  Carnegie Mellon University \\
  \texttt{ziangl@andrew.cmu.edu} \\
}

\begin{document}

\maketitle
\section{Problem Statement}
Typography plays an important role in publishing industry, and typically different typographies are needed for different use cases. However, designing a new font is very expensive and time-consuming, especially for Chinese, which involves a huge amount of characters. Some semi-automatic typography synthesis approaches have been proposed. These methods utilize structural commonalities to simplify this design process. One approach \cite{Chang:2017aa} is to design a subset of characters for new fonts manually first, then generate remaining characters automatically. In this project, we will try to explore this problem using Generative Adversarial Network (GAN).

\section{Related Work \& Motivation}
Traditional methods in Chinese typography synthesis view characters as an assembly of radicals and strokes, but they rely on manual definition of the key points, which is still time-costing. Some recent work on computer vision \cite{Isola:2016aa} proposes a brand new approach: to treat every Chinese character as an independent and inseparable image, so the pre-processing and post-processing of each character can be avoided. Then with a combination of a transfer network and a discriminating network, one typography can be well transferred to another as explained and demonstrated in \cite{Chang:2017aa}. Despite the quite satisfying performance of the model, the training process requires to be supervised, which means in the training data each character in the source domain and the target domain needs to be perfectly paired. Sometimes the pairing is time-costing, and sometimes there is no perfect pairing, such as the pairing between traditional Chinese and simplified Chinese characters.

\section{Proposed Approach}

\begin{figure}[h]
 \makebox[\textwidth]{\includegraphics[width=0.8\textwidth]{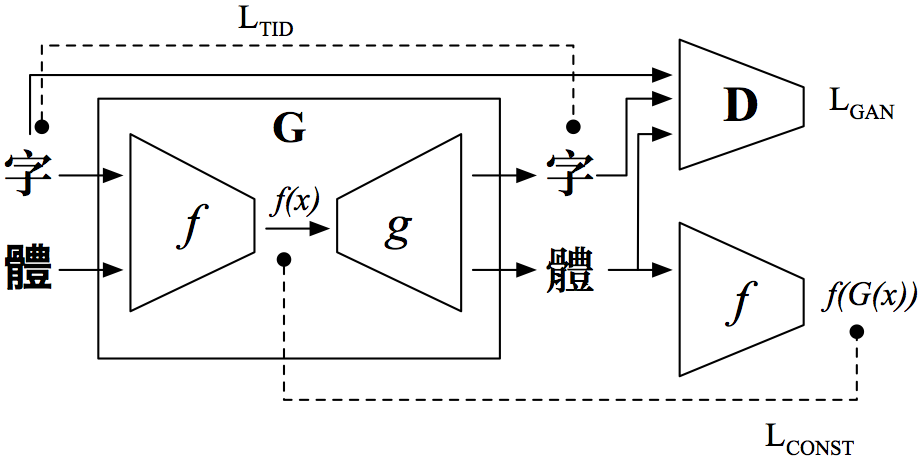}}
 \centering
 \caption{Architecture of our proposed method}
 \label{fig:arch}
\end{figure}

The input is a subset $T$ and a whole set $S$ of typography. Our proposed method transfers the style of $T$ onto $S$ and get a whole set of characters with style $T$.
Referring to other works on unsupervised style transferring \cite{Gatys_2016_CVPR, Taigman:2016aa,Zhu:2017aa}, we define three loss terms (fig \ref{fig:arch}). One term $L_{GAN}$ encourages to generate indistinguishable samples from the training samples of both source domain and target domain. The second term, $L_{CONST}$ is used for f-consistency after domain transferring, which means $f(x)$ and $f(G(c))$ should be close. The third term, $L_{TID}$, requires G to be close to the identity matrix for samples from target domain.

The $f$ function is usually a pretrained model consisting of a few convolutional layers that extract content representation of character images. The $g$ function is a series of deconvolutional layers adding target domain style to the character. We plan to use the pretrained encoder from zi2zi \footnote{\url{https://github.com/kaonashi-tyc/zi2zi/}}  project as $f$. We also propose to add skip connection trick introduced in U-Net to keep the edge of generated fonts sharp and avoid blurriness. 

\section{Data Description}

In the style transfer task, we use Noto Sans CJK as source and Noto Serif CJK as target\footnote{\url{https://www.google.com/get/noto/help/cjk/}} for the midway report. The former is a sans-serif font, while the latter is a serif one. 1000 pairs of characters are sampled from them: around 900 for training and 100 for testing. Each character is preprocessed into a $256 \times 256 \times 3$ vector. Currently our model has extended to a brush style font, SinoType XingKai\footnote{\url{http://zh.wikipedia.com/zh-hans/\%E5\%8D\%8E\%E6\%96\%87\%E8\%A1\%8C\%E6\%A5\%B7}}. Examples of the typographies can be seen in fig \ref{fig:data} of the appendix. Besides, we also generalized our model to some other Asian languages, including Japanese and Korean.

The pairs in the training set are not necessary to be perfectly aligned. For a training set with $N$ pairs, denote the source and target images in the $i$-th pair as $S_i$ and $T_i$, respectively. From strict to loose, we can define three pairing policies:

 \begin{enumerate}

\item \textbf{Strong Pair}: For any $i\in\{1,...,N\}$, $S_i$ and $T_i$ refer to the same character under different font style.

\item \textbf{Soft Pair}: For $i\in\{1,...,N\}$, $S_i$ and $T_i$ may not refer to the same character. However, for any $i$, there exists a value $j\in\{1,...,N\}$ so that $S_i$ and $T_j$ refer to the same character. In other words, though the element-wise pairing is wrong, the overlap between source font set and target font set is 100\% in the training set.

\item \textbf{Random Pair}: For $i\in\{1,...,N\}$, $S_i$ and $T_i$ may not refer to the same character. Moreover, for any $i$, there may or may not exist a value $j\in\{1,...,N\}$ so that $S_i$ and $T_j$ refer to the same character. 
 \end{enumerate}

\section{Model Description}
Our model uses the general architecture of zi2zi, which borrows the skip-connection trick from U-Net \cite{unet}. Both convolutional layers and deconvolutional layers use $2\times2$ strides and $5\times5$ kernels. Each convolutional layer is preceded by Leaky ReLu and followed by Batch Normalization. Each deconvolutional layer is followed by Batch Normalization and Skip Connection with corresponding convolutional layer. Deconv2 and Deconv3 are also followed by Dropout layer with 0.5 as dropout ratio. Deconv8 uses Conditional Instance Normalization\cite{CIN} instead of Batch Normalization. For the discriminator, instead of using binary classifiers, we use a trinary one introduced in DTN\cite{Taigman:2016aa}. The detailed architecture is shown in table \ref{tab configuration}.

\begin{table}[htbp]
\centering \caption{The structure of proposed method}
 \begin{tabular}{l|l}
  \toprule
  	\multicolumn{2}{l}{\textbf{Generator}} \\
  \midrule
    Conv-layers                             & Deconv-layers  \\
  \midrule
  Conv1: $256^{2}\times3\to128^{2}\times64$    & 
  Deconv1: $1^{2}\times(512+128)\to2^{2}\times1024$ \\
  Conv2: $128^{2}\times64\to64^{2}\times128$   & 
  Deconv2: $2^{2}\times(1024+512)\to4^{2}\times1024$ \\
                                                           \\
  Conv3: $64^{2}\times128\to32^{2}\times256$  & 
  Deconv3: $4^{2}\times(1024+512)\to8^{2}\times1024$ \\
    Conv4: $32^{2}\times256\to16^{2}\times512$ & 
    Deconv4: $8^{2}\times(1024+512)\to16^{2}\times1024$ \\
                                                           &  \\
    Conv5: $16^{2}\times512\to8^{2}\times512$ & 
    Deconv5: $16^{2}\times(1024+512)\to32^{2}\times512$ \\
    Conv6: $8^{2}\times512\to4^{2}\times512$  & 
    Deconv6: $32^{2}\times(512+256)\to64^{2}\times256$ \\
                                                         &  \\
  Conv7: $4^{2}\times512\to2^{2}\times512$   & 
  Deconv7: $64^{2}\times(256+128)\to128^{2}\times128$ \\
  Conv8: $2^{2}\times512\to1^{2}\times512$   & 
  Deconv8: $128^{2}\times(128+64)\to256^{2}\times3$ \\
         \midrule
	\multicolumn{2}{l}{\textbf{Discriminator}} \\
             \midrule

	\multicolumn{2}{l}{
    Conv: $256^{2}\times3\to
    128^2\times{64}\to
    64^2\times{128}\to
    32^2\times{256}\to
    16^2\times{512}$} \\

  	\multicolumn{2}{l}{
    FC: $16^2\times{512} \to 3$} \\

  \bottomrule
 \end{tabular}
\label{tab configuration}
\end{table}

Under Soft Pair policy, neither $L1$ nor $L2$ losses are available anymore. Therefore, one of our main modifications is adding losses $L_{TID}$, $L_{GAN}$ and $L_{CONST}$ to support unsupervised learning, as well as exploring further unsupervised mechanism: Random Pair. 

\newpage

\section{Experiments}

\subsection{Transfer performance under strong pair policy}

As a baseline, we implemented Chinese Typography Transfer (CTT) (\cite{Chang:2017aa}) and original zi2zi. And we implemented an unsupervised variant of zi2zi mentioned in Model Description. We built the training set under Strong Pair policy (900 pairs for training, 100 pairs for testing; Noto Sans CJK as source font, Noto Serif CJK as target font; 16 as minibatch size). All these three methods have decent performance and do not show significant difference.

\begin{figure}[h]
 \makebox[\textwidth]{\includegraphics[width=0.5\textwidth]{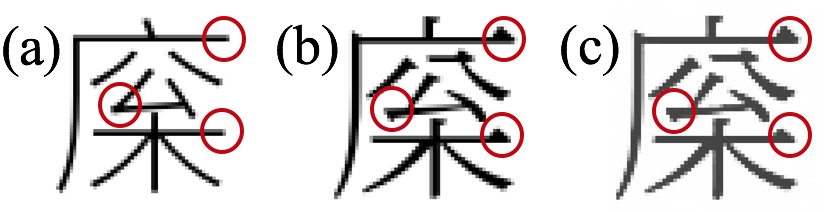}}
 \centering
    \vspace{-1em}
 \caption{Results using Strong Pair policy: (a) source font (b) target font (c) transferred font}
 \label{fig:han}
\end{figure}

The transfer result for CTT with Strong Pair is shown in fig \ref{fig:han}. Note that the serif feature is successfully transferred as indicated in the circles.

\subsection{Transfer performance under soft pair policy}

\begin{figure}[h]
   \makebox[\textwidth]{\includegraphics[width=1\textwidth]{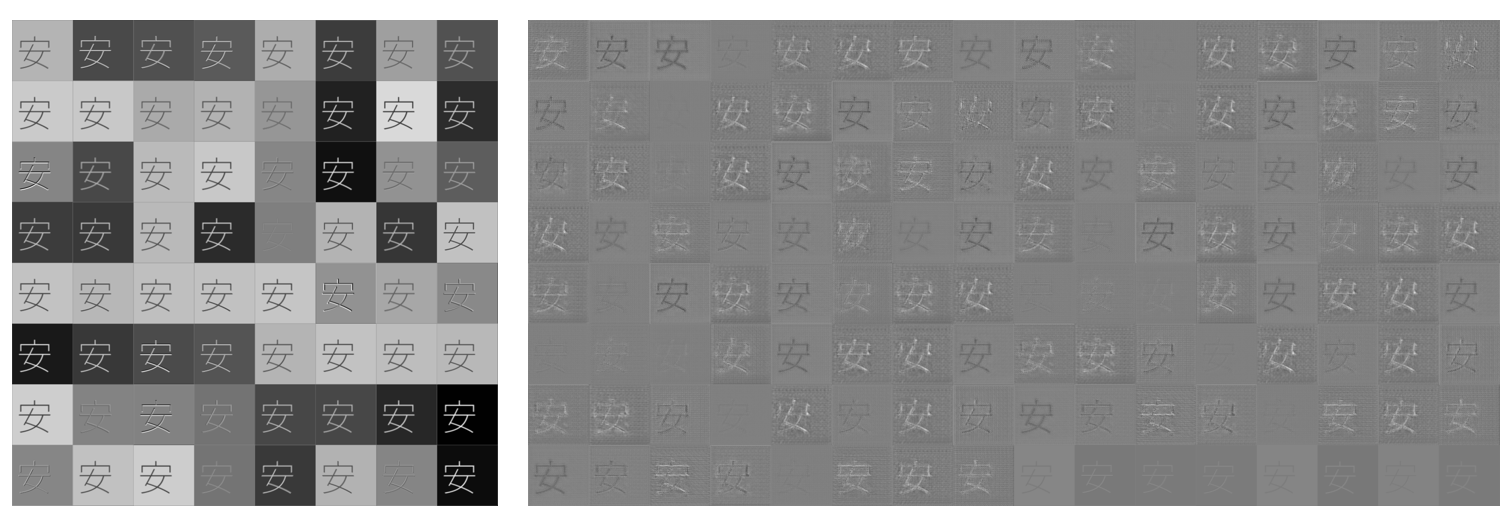}}
   \caption[caption]{Feature map of the model (left) convolution layer 1 (right) deconvolution layer 8}
   \label{fig:conv_and_deconv}
 \end{figure}

When switching to soft pair policy for training set generation, with $L2$ loss between transfer font and target font disabled, none of existing approaches work well. The transferred fonts are unstable, and even converge into messy black blocks. \\
However, our method still works well under soft pair policy. It converges faster than zi2zi and CTT. After 5 epochs, we observe blurry serifs. The serifs become clear and sharp after 12 epochs. Fig \ref{fig:conv_and_deconv} shows the feature map of th convolution layer 1 and deconvolution layer 8, respectively.

\subsection{Transfer performance under random pair policy}

\begin{figure}[h]
	\begin{center}
		\begin{minipage}{0.16\textwidth}
			\includegraphics[width=\linewidth]{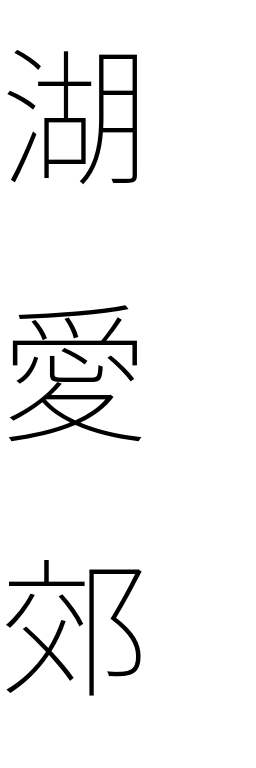}
		\end{minipage}
		\begin{minipage}{0.16\textwidth}
			\includegraphics[width=\linewidth]{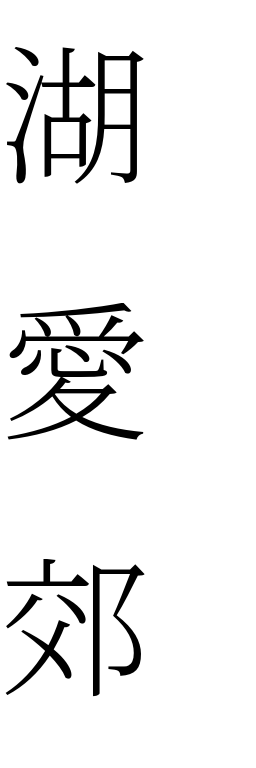}
		\end{minipage}
		\begin{minipage}{0.16\textwidth}
			\includegraphics[width=\linewidth]{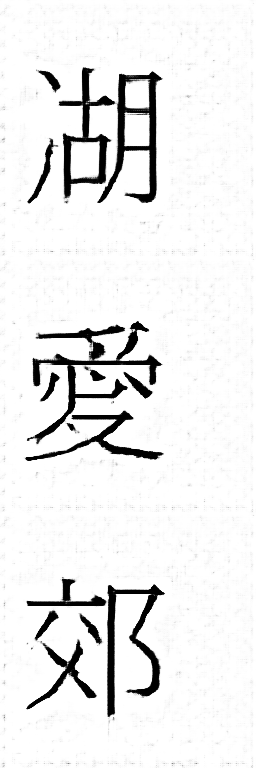}
		\end{minipage}
		\begin{minipage}{0.16\textwidth}
			\includegraphics[width=\linewidth]{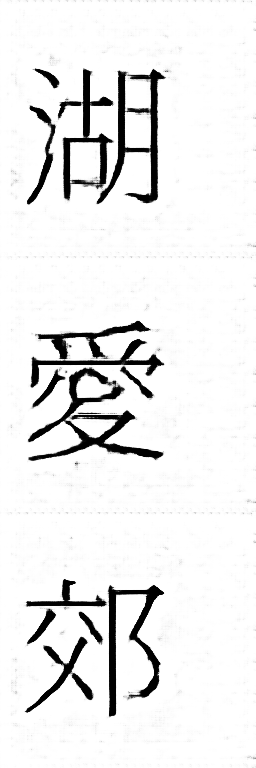}
		\end{minipage}
		\begin{minipage}{0.16\textwidth}
			\includegraphics[width=\linewidth]{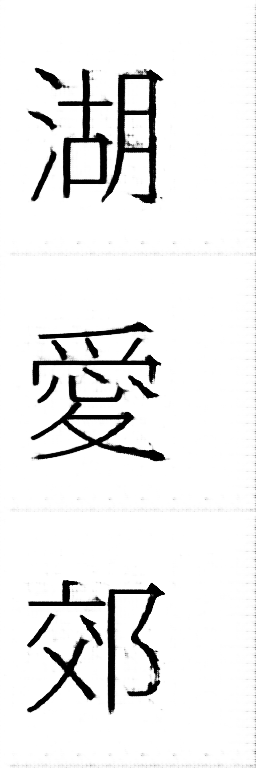}
		\end{minipage}
    \end{center}
    \caption{Effect of overlap ratio. Comparison of the generated results after 100 epochs. Column 1: Source font. Column 2: Ground truth of target font. Column 3: Overlap ratio = 0. Column 4: Overlap ratio = 0.5. Column 5: Overlap ratio = 1.}
    \label{fig:overlap}
\end{figure}

We explored the model performance under three overlap ratios (0.0, 0.5 and 1.0). The model was able to learn the serif feature in all the three settings. However, the model performs poorly under a low overlap ratio in two ways. First, the background of inferred images is more noisy. Also, the structure of characters may be altered in the inferred images. For example, some strokes may be lost. The differences are illustrated in fig \ref{fig:overlap}.

\subsection{Transfer performance on calligraphy fonts} 

\begin{figure}[h]
   \makebox[\textwidth]{\includegraphics[width=1\textwidth]{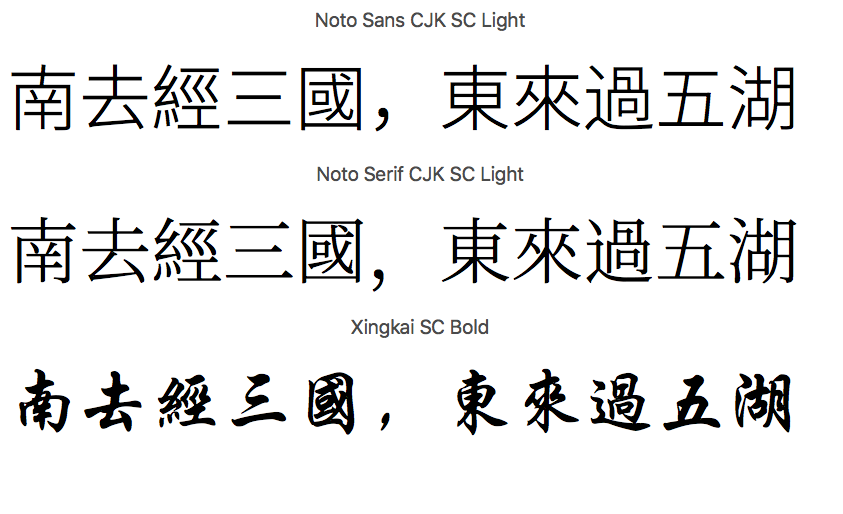}}
    \makebox[\textwidth]{\includegraphics[width=1\textwidth]{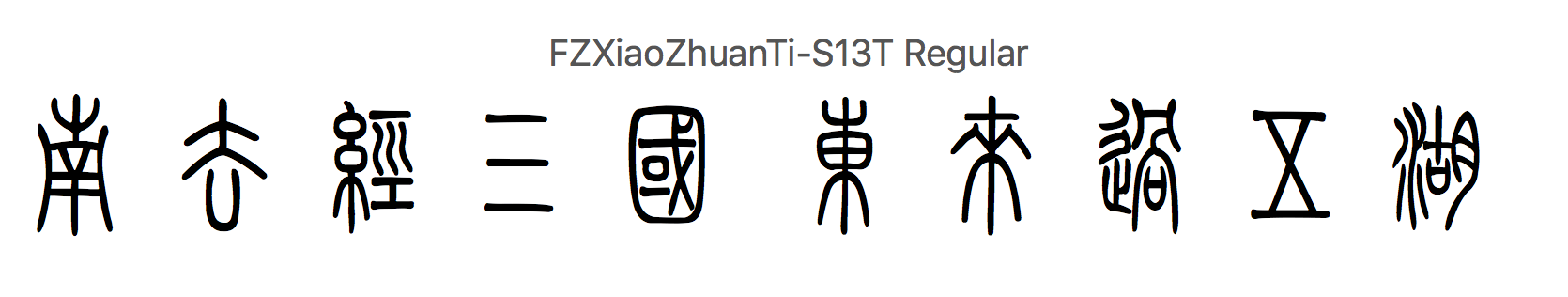}}
   \caption{Typographies used in our project}
   \label{fig:data}
 \end{figure}

Calligraphy font is a font family very different from sans-serif and serif ones. Style transfer between non-calligraphy and calligraphy font looks more difficult due to the significant difference in font skeleton, as can be seen in fig \ref{fig:data}. 

We did experiment by transferring the style of SinoType XingKai onto Noto Sans CJK, with soft pair policy or zero-overlap random pair policy (fig \ref{fig:XingKai_soft} and fig \ref{fig:XingKai_random}). It is difficult to tell whether soft pair policy or random pair policy leads to better performance. We also observed the convergence is much slower than previous experiments. 





 \begin{figure}[h]
   \makebox[\textwidth]{\includegraphics[width=0.3\textwidth]{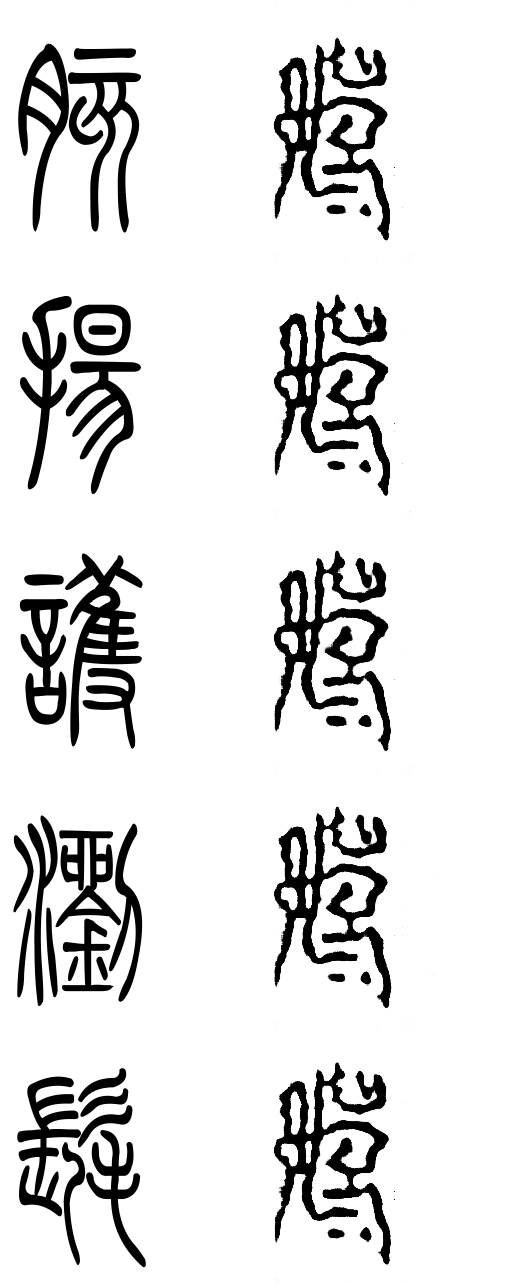}}
   \caption[caption]{The mode missing issue. (Left) characters in target font (Right) characters in generated font}
   \label{fig:mode_missing}
 \end{figure}

 In another experiment, an ancient Chinese font, FZ Xiaozhuan, is tried as target font. We encountered a common issue of GANs called ``model missing issue'' \cite{modemissing}, in which results lost diversity and generated characters are all similar and unrecognizable (see fig \ref{fig:mode_missing}). We solved this issue by adding random shifting and scaling for source and target characters in each training iteration. The random shift and scaling is used as both data augmentation as well as regularization. Using this trick, the result looks much better, as can be seen in fig \ref{fig:XiaoZhuan_soft}.

 
\subsection{Transfer performance for non-Chinese languages} 

\begin{figure}[h]
   \makebox[\textwidth]{\includegraphics[width=0.8\textwidth]{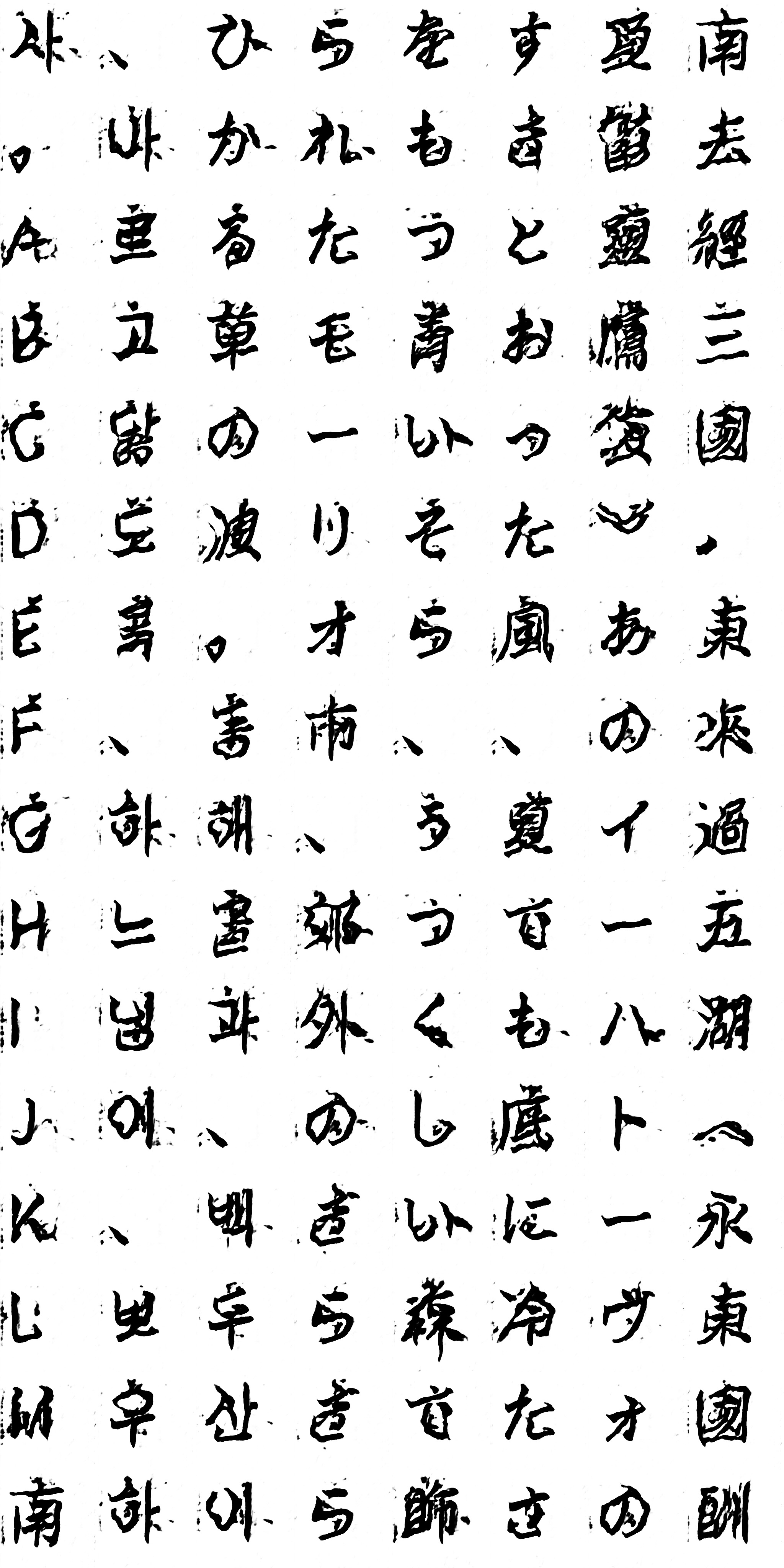}}
   \caption[caption]{Transferred typography for Chinese, Japanese, Korean and English characters with soft pair policy (Noto Sans CJK as source font, SinoType Xingkai as target font) }
   \label{fig:XingKai_soft}
 \end{figure}

\begin{figure}[h]
   \makebox[\textwidth]{\includegraphics[width=0.8\textwidth]{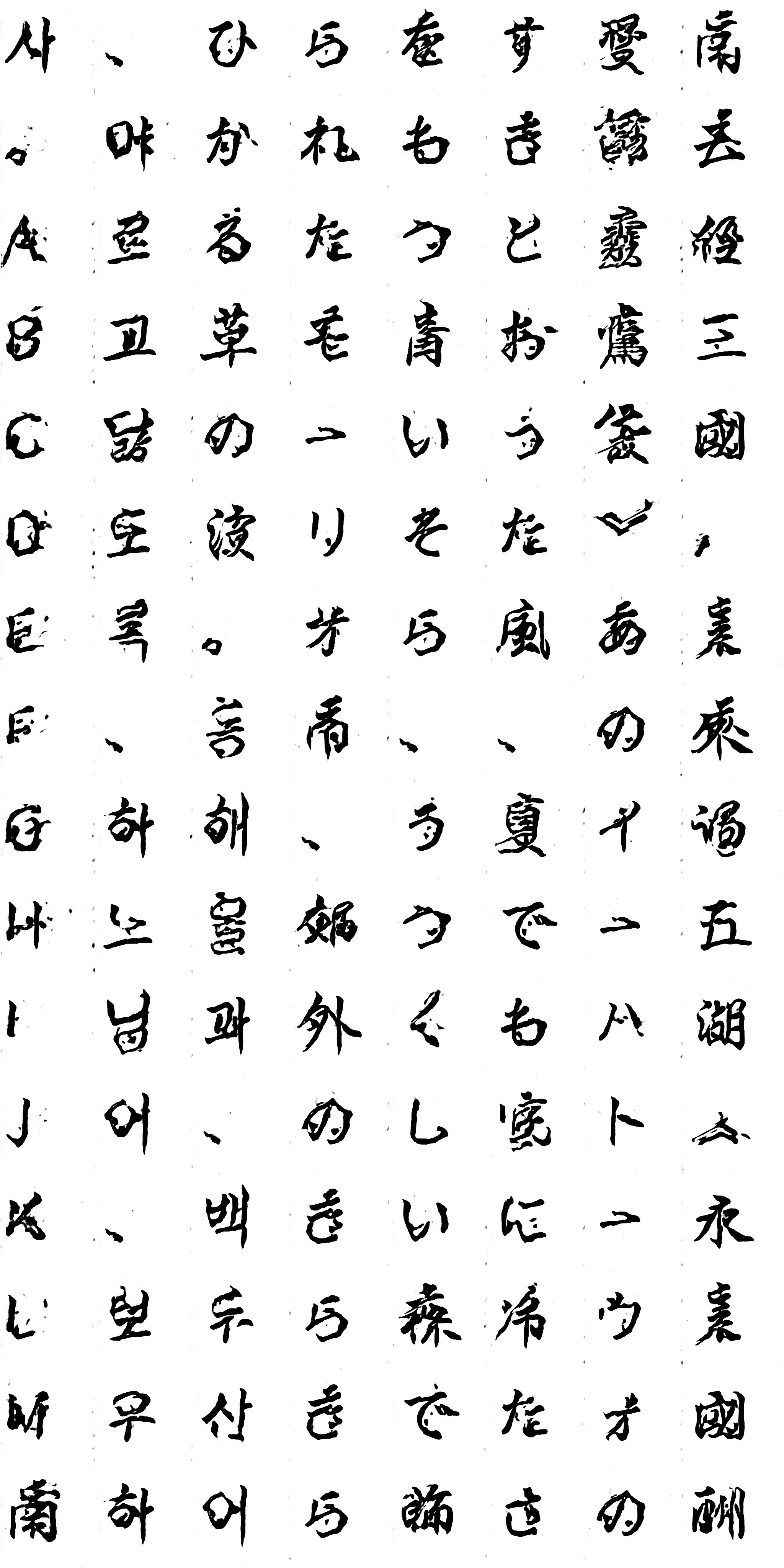}}
   \caption[caption]{Transferred typography for Chinese, Japanese, Korean and English characters with zero-overlap policy (Noto Sans CJK as source font, SinoType Xingkai as target font) }
   \label{fig:XingKai_random}
 \end{figure}

\begin{figure}[h]
   \makebox[\textwidth]{\includegraphics[width=0.8\textwidth]{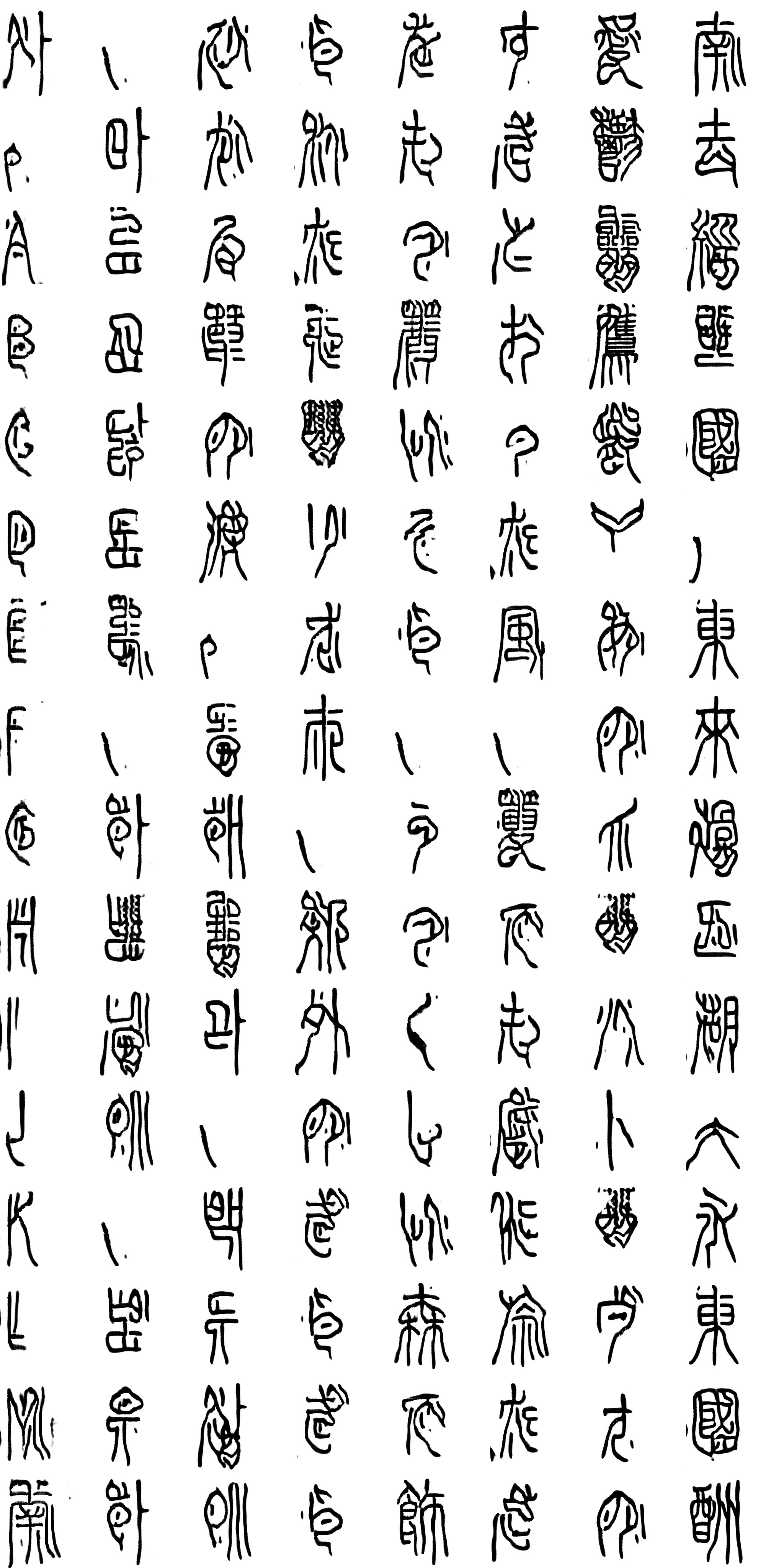}}
   \caption[caption]{Transferred typography for Chinese, Japanese, Korean and English characters with soft pair policy (Noto Sans CJK as source font, FZ Xiaozhuan as target font) }
   \label{fig:XiaoZhuan_soft}
 \end{figure}

Though we only use simplified Chinese characters in the training period, the model could still generalize on traditional Chinese characters and even non-Chinese ones. In figure \ref{fig:XingKai_soft}, \ref{fig:XingKai_random} and \ref{fig:XiaoZhuan_soft}, the rightmost column are traditional Chinese characters. The leftmost column are Latin characters. Other columns are a mixture of Japanese and Korean characters.

\subsection{Effect of various loss weights}

\begin{figure}[h]
   \makebox[\textwidth]{\includegraphics[width=0.3\textwidth]{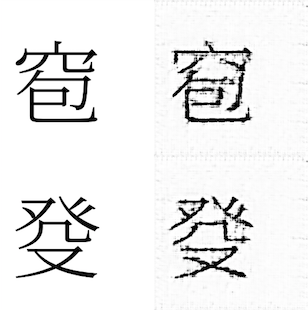}}
   \caption[caption]{Transferred result on Noto Sans CJK and Noto Serif CJK with $L_{TID}=0$ (left) ground-truth source font (right) transferred font}
   \label{fig:zero_Ltid}
 \end{figure}

We implemented the mechanism to assign the weight of $L_{TID}$, $L_{CONST}$ and $L_{TV}$ to evaluate the effect of these losses. \\
In the midway report, we mentioned $L_{TID}$ loss is quite crucial to the performance. This is better illustrated by an extreme case: manually assigning 0 to $L_{TID}$ and the result between Noto Sans CJK and Noto Serif CJK is shown in fig \ref{fig:zero_Ltid}.

As can observed from fig \ref{fig:zero_Ltid}, after 100 epochs the serif-transferring performance is poor and there are still noisy pixels in the background. By setting other weights to $L_{TID}$, it can be observed that the serif-transferring effect on the same two fonts gets significantly better as the weight of $L_{TID}$ increases (but also need to balance in cooperation with other losses). Therefore in the final default setting, $L_{TID}$ is assigned with a considerably larger weight than the other losses.

\begin{figure}[h]
	\begin{center}
		\begin{minipage}{0.18\textwidth}
			\includegraphics[width=\linewidth]{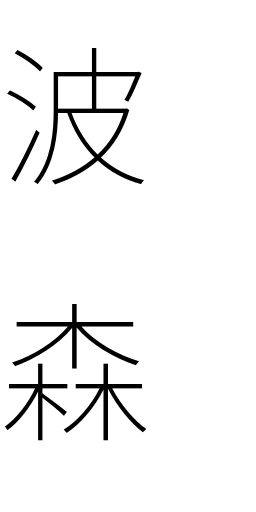}
		\end{minipage}
		\begin{minipage}{0.18\textwidth}
            \includegraphics[width=\linewidth]{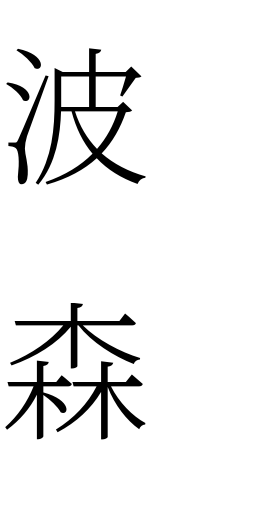}
        \end{minipage}
		\begin{minipage}{0.18\textwidth}
            \includegraphics[width=\linewidth]{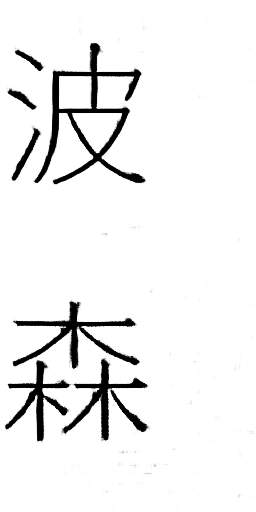}
        \end{minipage}
		\begin{minipage}{0.18\textwidth}
	        \includegraphics[width=\linewidth]{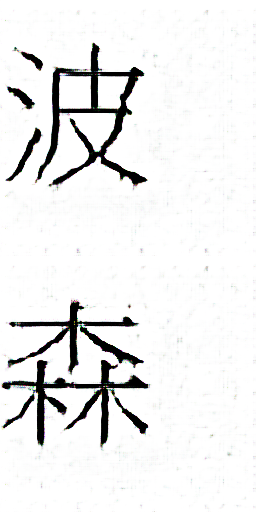}
        \end{minipage}
	\end{center}
    \caption{Effect of $L_{CONST}$. Comparison of the generated results after 100 epochs. Column 1: Source font. Column 2: Ground truth of target font. Column 3: Without $L_{CONST}$. Column 4: With $L_{CONST}$.}
    \label{fig:lconst}
\end{figure}

For $L_{CONST}$, we compared the results when training with and without $L_{CONST}$ loss. Without the loss, the inferred images have cleaner strokes and background. However, the serif feature is missing or not obvious on some characters, although quite obvious on some others. Compared with the results with $L_{CONST}$ loss, the inferred images are closer to the source topography. This is illustrated in fig \ref{fig:lconst}.

\subsection{Effect of pretraining}



In the midway report, we mentioned that pretraining did not help to improve performance. However, we later found that there was an error in the experiment process. We fixed this error and did the experiment again.

\begin{figure}[h]
\begin{center}
	\begin{minipage}{0.18\textwidth}
		\includegraphics[width=\linewidth]{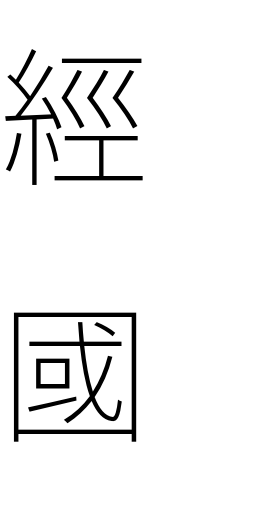}
	\end{minipage}
	\begin{minipage}{0.18\textwidth}
		\includegraphics[width=\linewidth]{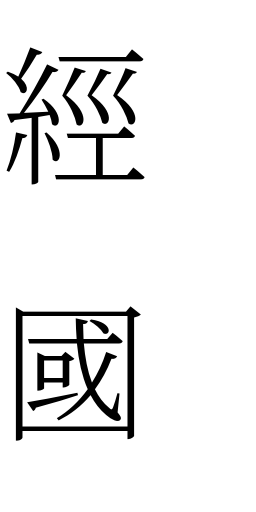}
	\end{minipage}
	\begin{minipage}{0.18\textwidth}
		\includegraphics[width=\linewidth]{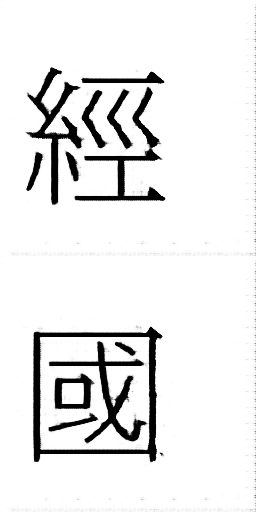}
	\end{minipage}
	\begin{minipage}{0.18\textwidth}
		\includegraphics[width=\linewidth]{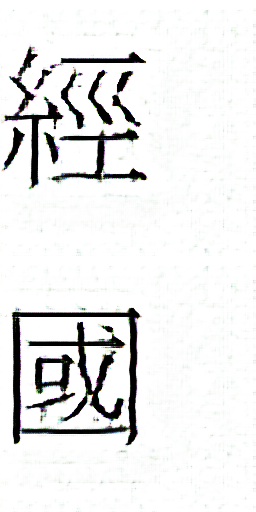}
	\end{minipage}
\end{center}
\caption{Effect of pretraining. Comparison of the generated results after 100 epochs. Column 1: Source font. Column 2: Ground truth of target font. Column 3: Generated with pretraining. Column 4: Generated without pretraining.}
\label{fig:pretrain}
\end{figure}

The inferred images are much better when the model is trained using pretrained encoder from zi2zi.  With pretraining, the strokes are stright and clean, whereas without pretraining, strokes are sometimes torn into pieces, and a straight stroke can sometimes be curved. The ``serif'' feature is also cleaner with pretraining. These are shown in fig \ref{fig:pretrain}.

With pretraining, the model behavior in the training process is also more stable. Due to randomness in Stochastic Gradient Descent, the model parameters can sometimes reach a poor setting, and the inferred images can be very noisy. This phenomenon is worse without pretraining than with pretraining.

\subsection{Effect of batch normalization}

\begin{figure}[h]
   \makebox[\textwidth]{\includegraphics[width=0.9\textwidth]{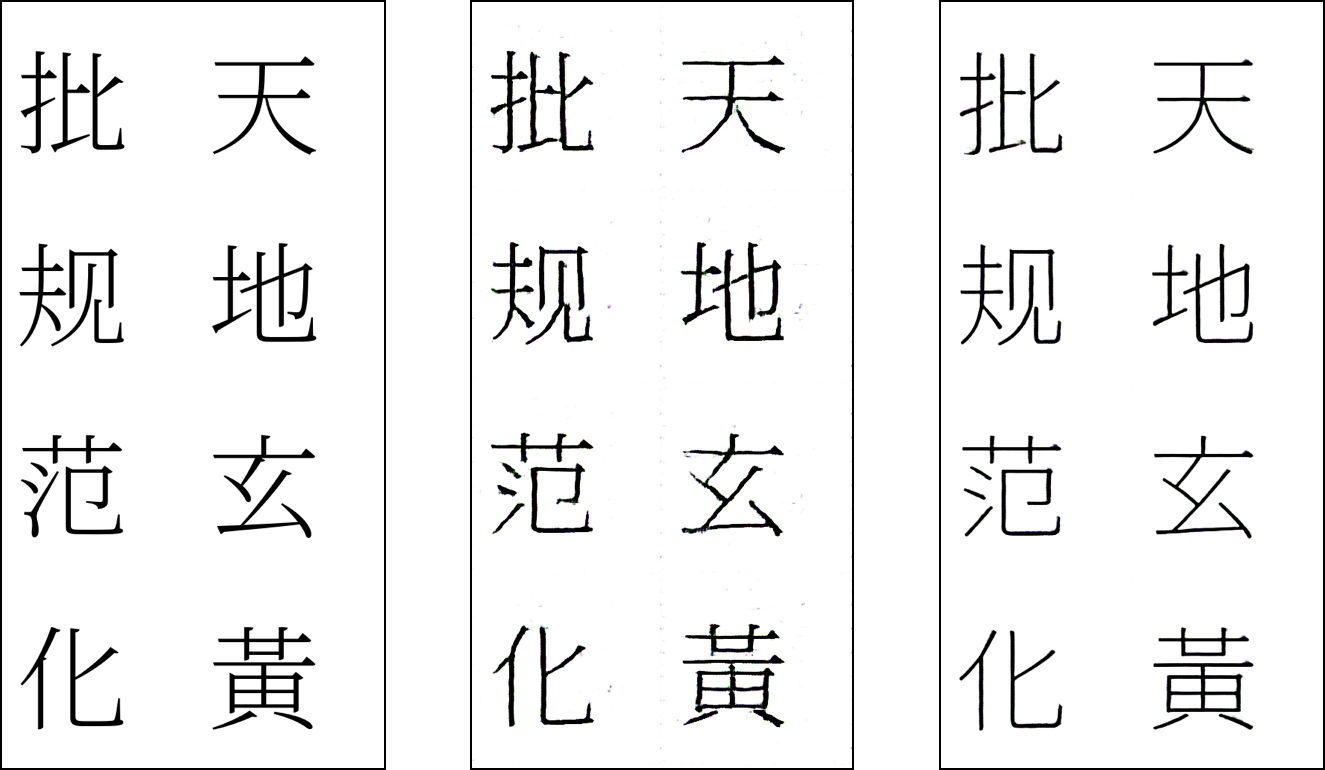}}
   \caption[caption]{The effect of phase flag for Batch Normalization on transferred font\\\hspace{\textwidth}(left) ground truth (middle) training phase (right) inference phase}
   \label{fig:bn}
 \end{figure}

Batch normalization cell has two phases, training phase and inference phase. Different phases have different effects on the generated result. We found that in the training phase, the generated characters are more noisy and have sharp serifs. In the inference phase, the generated ones are less noisy but lose the sharp serifs as well as some similarity with target font, as can be seen in figure \ref{fig:bn}. While Conditional Instance Normalization is already used to alleviate this issue, we will continue to explore methods to get a trade-off between noise and similarity.

\section{Evaluation}

Currently, we used $L2$ loss between generated character and true target character, and empirical rating to evaluate the model's performance. However, $L2$ loss may not be a good estimation for the quality of generated typography. We will try to design turning tests as another qualitative and quantitative measurement. We will shuffle generated result and ground-truth result and ask volunteers to separate them out. If the separation task is difficult for humans, we can show that our model's performance is good enough.

\section{Future Work}

Our current model uses Convolutional/Deconvolutional neural units for style transfer. They are based on the locality assumption and may not have a holistic view of the source/target typography. We are curious about whether attention mechanism \cite{attention} can further improve model's performance by utilizing some long-distance image context. 

As mentioned before, the architecture we use may suffer the ``mode missing issue''. Existing losses, including $L2$ loss, $L_{CONST}$ and $L_{TID}$, can't detect when the SGD is trapped into this local optimum efficiently. Inspired by \cite{gancreatedequal}, we could adapt Frechet Inception Distance to the typography style transfer domain as a metric of model robustness and result diversity.

\section{Acknowledgement}

We want to thank Jie Chang especially, who is the author of \cite{Chang:2017aa}, for his helpful discussion with us about this project.

\bibliography{proposal}
\newpage
\section*{Appendix}

   \begin{figure}[h]
   \makebox[\textwidth]{\includegraphics[width=0.8\textwidth]{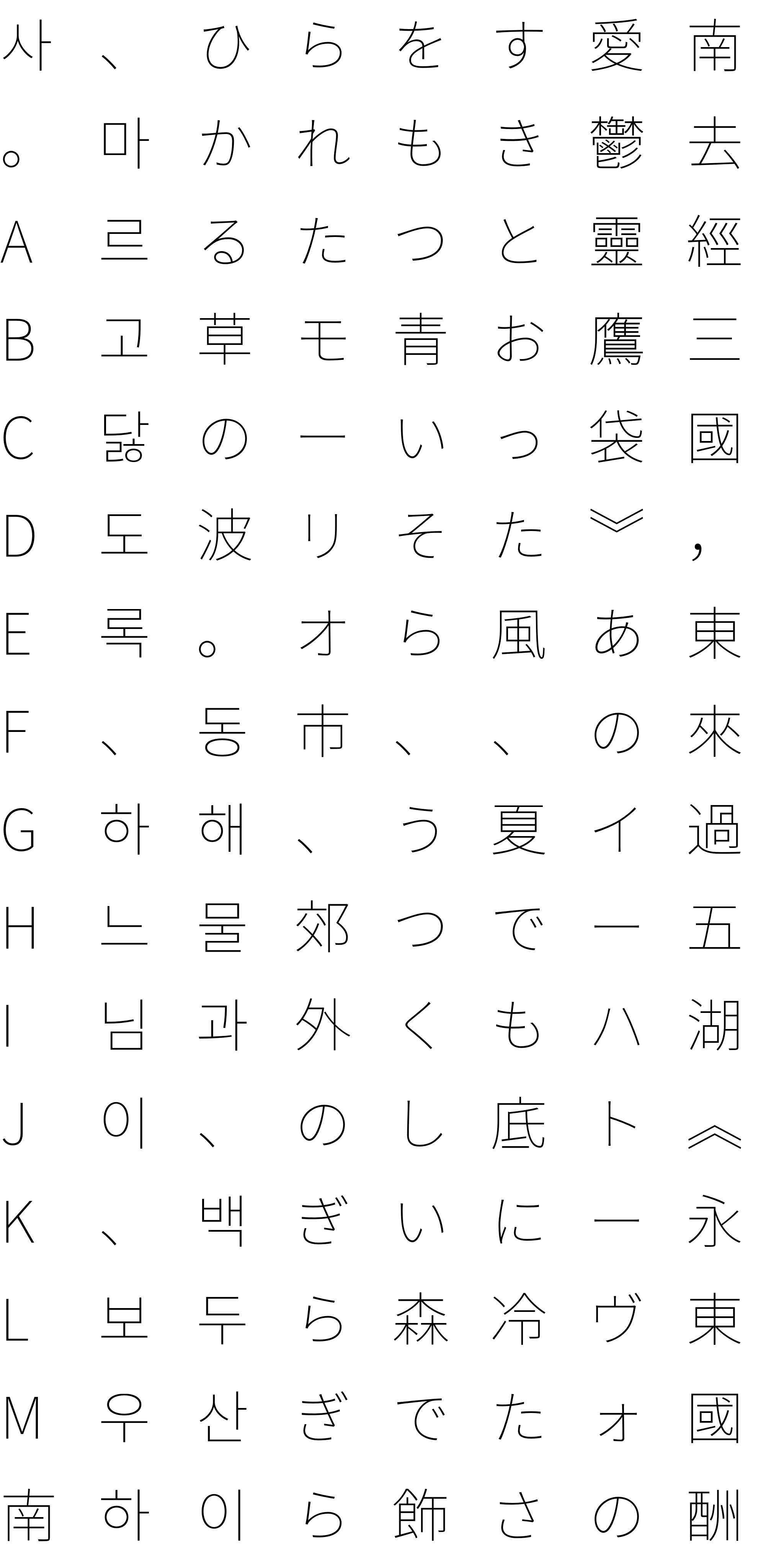}}
   \caption[caption]{Source font (Noto Sans CJK) used for style transfer }
   \label{fig:SourceFont}
 \end{figure}


\end{document}